\PassOptionsToPackage{unicode}{hyperref}
\PassOptionsToPackage{hyphens}{url}
\documentclass[
]{article}
\usepackage{xcolor}
\usepackage{amsmath,amssymb}
\usepackage{tipa}
\usepackage{iftex}
\ifPDFTeX
  \usepackage[T1]{fontenc}
\newfontfamily\garamond{EBGaramond}[
  Extension = .otf,
  UprightFont = *-Regular,
  ItalicFont = *-Italic,
  BoldFont = *-Bold ]
  \usepackage[utf8]{inputenc}
  \usepackage{textcomp} % provide euro and other symbols
\else % if luatex or xetex
  \usepackage{unicode-math} % this also loads fontspec
  \defaultfontfeatures{Scale=MatchLowercase}
  \defaultfontfeatures[\rmfamily]{Ligatures=TeX,Scale=1}
\fi
\usepackage{lmodern}
\ifPDFTeX\else
\fi
\IfFileExists{upquote.sty}{\usepackage{upquote}}{}
\IfFileExists{microtype.sty}{% use microtype if available
  \usepackage[]{microtype}
  \UseMicrotypeSet[protrusion]{basicmath} % disable protrusion for tt fonts
}{}
\makeatletter
\@ifundefined{KOMAClassName}{% if non-KOMA class
  \IfFileExists{parskip.sty}{%
    \usepackage{parskip}
  }{% else
    \setlength{\parindent}{0pt}
    \setlength{\parskip}{6pt plus 2pt minus 1pt}}
}{% if KOMA class
  \KOMAoptions{parskip=half}}
\makeatother
\usepackage{graphicx}
\makeatletter
\newsavebox\pandoc@box
\newcommand*\pandocbounded[1]{% scales image to fit in text height/width
  \sbox\pandoc@box{#1}%
  \Gscale@div\@tempa{\textheight}{\dimexpr\ht\pandoc@box+\dp\pandoc@box\relax}%
  \Gscale@div\@tempb{\linewidth}{\wd\pandoc@box}%
  \ifdim\@tempb\p@<\@tempa\p@\let\@tempa\@tempb\fi% select the smaller of both
  \ifdim\@tempa\p@<\p@\scalebox{\@tempa}{\usebox\pandoc@box}%
  \else\usebox{\pandoc@box}%
  \fi%
}
\def\fps@figure{htbp}
\makeatother
\ifLuaTeX
  \usepackage{luacolor}
  \usepackage[soul]{lua-ul}
\else
  \usepackage{soul}
\fi
\usepackage{bookmark}
\IfFileExists{xurl.sty}{\usepackage{xurl}}{} % add URL line breaks if available
\hypersetup{
  hidelinks,
  pdfcreator={LaTeX via pandoc}}

\author{}
\date{}
\newfontfamily\garamond{EBGaramond}[
  Extension = .otf,
  UprightFont = *-Regular,
  ItalicFont = *-Italic,
  BoldFont = *-Bold ]
\newcommand{\subhead}[1]{\par\vspace{0.8em}\noindent\textbf{#1}\par\nopagebreak}
\usepackage{xurl}
\newcommand{\ced}[1]{\ooalign{#1\cr\hidewidth\char"00B8\hidewidth\cr}}
\begin{document}

\textbf{
\section{Yorùbá in Unicode: An Overview of a Problem}
}

\emph{By Kọ́lá Túbọ̀sún}

Yoruba Names Project (project@yorubaname.com \textbar{}
www.YorubaName.com)

\vspace{1em}
\noindent\rule{\linewidth}{0.4pt}
\vspace{1em}

\emph{Abstract:}

There's a recurrent problem in the writing of Yorùbá on the internet and
on the computer that has proven intractable over the years. The language
(along with other African languages that depend on diacritics for
disambiguation) requires a small set of precomposed characters that
Unicode does not encode. This has forced writers and digital systems to
rely on combining character sequences that behave inconsistently across
platforms, corrupt under font substitution, and fail in search. This
paper documents that failure across a range of real-world contexts, from
published books to web platforms to mobile keyboards, using personal and
empirical evidence. It identifies Unicode\textquotesingle s NFC
normalization stability policy as the structural constraint that
prevents a straightforward fix, arguing for direct intervention of the
Consortium in solving the active problem, proposing a formal encoding
request for the four core Yorùbá characters as the most durable path to
resolution.

\subhead{Yorùbá Tonal Ambiguities}

Yorùbá, a Niger-Congo language spoken by over 40 million speakers around
the world, is a tone language (Bámgbóṣé, 1966; Akinlabí, 2004), which
means that pitch height in the language is significant in meaning
change. This is different from intonational languages in which pitch
variation (intonation) is significant for grammatical function or mood.
So, in English, for instance, ``You have come'' and ``You have come?''
sound different because of the intonation placed on the last word. It
changes the meaning of the sentence from a statement to a question, but
does not quite do anything to the meaning of the word ``come'' on which
the intonation is set. About sixty to seventy percent of world languages
are tonal (Yip, 2002; Pulleyblank, 2004).

In Yorùbá, the difference between ọwọ́, ọwọ̀ and ọ̀wọ̀ are lexically
significant. They mean \textquotesingle hand\textquotesingle,
\textquotesingle broom\textquotesingle, and
\textquotesingle respect\textquotesingle{} respectively. (When written
with a capital Ọ̀ the last one, Ọ̀wọ̀ means a town in Ondo State of
Nigeria).

The presence of this type of lexical ambiguity made it important that
the orthography of the language be capable of providing proper
disambiguation when the words are written. Otherwise, a sentence like
``Fun mi l'owo'' without any markings can mean anything from `Give me a
hand' to `Give me respect'. Longer sentences provide even longer issues
for comprehension, as I showed while teasing out at least fifty-five
English {language meaning possibilities out of a simple Yorùbá
sentence presented without diacritics: ``Baba mi ni oko nla.''}

Here are the first ten:

\begin{enumerate}
\def\labelenumi{\arabic{enumi}.}
\item
  Father moved inside the big farm (bàbá mì ní oko ńlá)
\item
  Father moved/shook inside the big car (bàbá mì ní ọkọ̀ ńlá)
\item
  Sorghum shook inside the big farm (bàbà mì ní oko ńlá)
\item
  Sorghum shook inside the big car (bàbà mì ní ọkọ̀ ńlá)
\item
  My bronze is a big car (bàbà mi ni ọkọ̀ ńlá)
\item
  My bronze has a big car (bàbà mi ní ọkọ̀ ńlá)
\item
  My bronze is a big farm (bàbà mi ni oko ńlá)
\item
  My bronze has a big farm (bàbà mi ní oko ńlá)
\item
  My bronze is a big penis (bàbà mi ni okó ńlá)
\item
  My bronze has a big penis (bàbà mi ní okó ńlá)
\end{enumerate}

(Túbọ̀sún, 2018a)

The difficulty these ambiguities pose for non-native learners has been
documented experimentally: Ọlátúbọ̀sún (2012) found that adult American
English speakers consistently struggled with tonal contrast in
production even when their pitch range fell within native speaker norms,
suggesting that the problem is one of relative pitch discrimination
rather than absolute pitch capacity.

\subhead{A Short History of Yorùbá Orthography}

Yorùbá was not always a written language, and there is yet no evidence
that it had an indigenous script (Brandon, 2008). It was not until
colonial times that writing developed in Yorubaland ({Oyěníyì},
2016).\footnote{Author names in this paper are marked for tone in full
  throughout this paper, in both body and references, including contour
  tones as is this particular case, following the notation argued for in
  Túbọ̀sún (2026). So instead of unmarked names like Ogunbiyi,
  Oyeniyi, Oyetade, etc, or partially-marked contours like Ògúnbíyì,
  Oyèníyì, Oyètádé, etc more common in literature, I give instead the
  fully marked forms: Ògúnbíyǐ, Oyěníyì, Oyětádé. A reader searching for
  these names, of course, will still need to use the most common forms.
  That this paper cannot cite its own sources in full orthography
  without rendering them harder to find \emph{is} one of the arguments
  this paper makes, reproduced in the bibliography.} But shortly before
contact with Europeans in West Africa, Yorùbá had been written with a
system called Ajami, based on Arabic script. They had traded with the
Arabs through the trade routes in the desert. But the arrival of British
contact, slave trade, missionary activity, and eventually colonialism,
eroded the development of Ajami, and replaced it with the Roman script.

The presence of tone in Yorùbá, however, made the adoption of the Roman
script a more demanding undertaking, just as it did in Ajami, as
Ògúnbíyǐ (2003) documents. The early writers of the language quickly
found out that unlike what obtains in English, one would need extra
character modifications on the Roman script to properly write in
intelligible Yorùbá. This was not an easy decision to come to. Àjàyí
{(1960) describes those early days of missionary activity in Nigeria
where Reverend Samuel} Àjàyí {Crowther, then a newly ordained
minister of the Church Missionary Society (C.M.S), and Rev. C.A.
Gollmer, a missionary at Badagry who was originally from Germany, first
broached the idea of writing with diacritics. Gollmer, who had
experience with other European languages like Italian, French, Latin,
Greek, and Hebrew, thought that it was better to model the writing after
Italian (where ``one sound should be represented by one letter only'')
rather than the English model with its irregular spellings, where} the
\textquotesingle ou\textquotesingle{} in the words
\textquotesingle through\textquotesingle,
\textquotesingle thought\textquotesingle{} and
\textquotesingle though\textquotesingle, for instance, are pronounced
differently.

But the reactions to this suggestion reportedly ranged from tepid to
negative to hostile, with a certain Prof. Lee quoted as having written
that apart from accents, ``I would use no other diacritical mark
whatsoever\ldots{} and would rather use double or combined
letters\ldots{} perhaps such marks as the French cedille {ç} or the
Polish \emph{ą} may be excepted'' (Àjàyí, 1960, p. 51). The author
however allowed for the probability that the objections could have also
come from the anticipated problems of the printing press at the C.M.S,
justifying the preference of the \emph{aw} over the \emph{ọ}.
Eventually, even when some of these modest diacritical markings were
accepted, they were still not universally accepted, such that when the
Prayer Book was published in 1850, it was absent of diacritics
completely. Rev. W. Knight, the C.M.S. Secretary justified this in a
December 1852 letter to Crowther implying that they might become
obstacles to ``a native's acquiring the art of reading'' since ``to
teach them to read is our great aim'' (ibid, p. 53)

While this was going on, books were being published that utilized
diacritics to different levels of (in)consistency. Rev. Crowther had
published his \emph{The Vocabulary of Yorùbá} in 1843. It was, except
from the short word lists by his earlier European missionary mentors,
the first major work of Yorùbá lexicography, using a system that neither
fully used diacritics nor totally avoided them. The influence of the
English-style orthography supporters was evident in the irregular
spellings that would leave today's reader very confused. For instance, a
word for `truly', which will be written today as `l'óòótọ́' was written
by Crowther as `Lotoh' (page 31). And to use the example of `owo'
earlier referenced, these spellings were found:

\begin{quote}
``Ohwoh'' (\emph{hand})

``Ohwor'' (\emph{broom}/\emph{besom})

``Owo/owó'' (\emph{money})

``Owo'' (\emph{boil}).
\end{quote}

\emph{Source: The Vocabulary of Yorùbá (Crowther, 1843).}

In modern Yorùbá, the last word there, for boil, would be written as
`oówo'. What was evident from Crowther's early work, even from this
small sample example, was the absence of a recognizable system. Nothing
in the text could have told a reader why the word for `hand' and the
word for `broom' differed only in one letter \emph{h}/\emph{r} even
though their difference was one of tone: \emph{Ọwọ́} (hand): \emph{Ọwọ̀}
(broom). Other examples with other words in the language showed this
more clearly.

\begin{quote}
``Ehhin'' (back)

``Ehin/Eyin'' (tooth)

``Ehnyi'' (You (pl))

``Ehyin'' (egg).\footnote{ibid.}
\end{quote}

The closest comparison here would be the word for `you' and the one for
`back'. In modern Yorùbá, that would be `ẹ̀yin' and `ẹ̀yìn' respectively
--- also a matter of tone, the former with a mid tone on the second
syllable while the latter with a low tone in the same position. Nothing
in Crowther's option was consistent with the decision he made earlier
with `broom' and `hand'. And so, \emph{The Vocabulary}, though very
useful as a first attempt at grappling with the complexity of Yorùbá
speech in writing, did little to help readers come to an intuitive
understanding of Yorùbá tone behaviour.

But by the middle of the 19th century, more Africans had started writing
and publishing in the language, each of them influenced by their own
German or French mentors, personal opinion of how the language should be
written, and mistakes they have noticed in Rev. Crowther's work. This
eventually led to the Conference of 1875 where some basic rules were
set, and with which Yorùbá was written into the 20th century. It
includes the adoption of `gb' (and not \emph{bh} or \emph{b}) for the
labial-velar plosive, `p' instead of `kp', subdots \emph{ọ}, \emph{ẹ},
\emph{ṣ}, the tilde \textasciitilde{} on nasalized vowels and on contour
tones, and the curious rule to use the tonal diacritic only on the first
syllable of the ``polysyllabic word carrying the same tone throughout''
(page 55), among others.

\includegraphics[width=5.36363in,height=4.16666in]{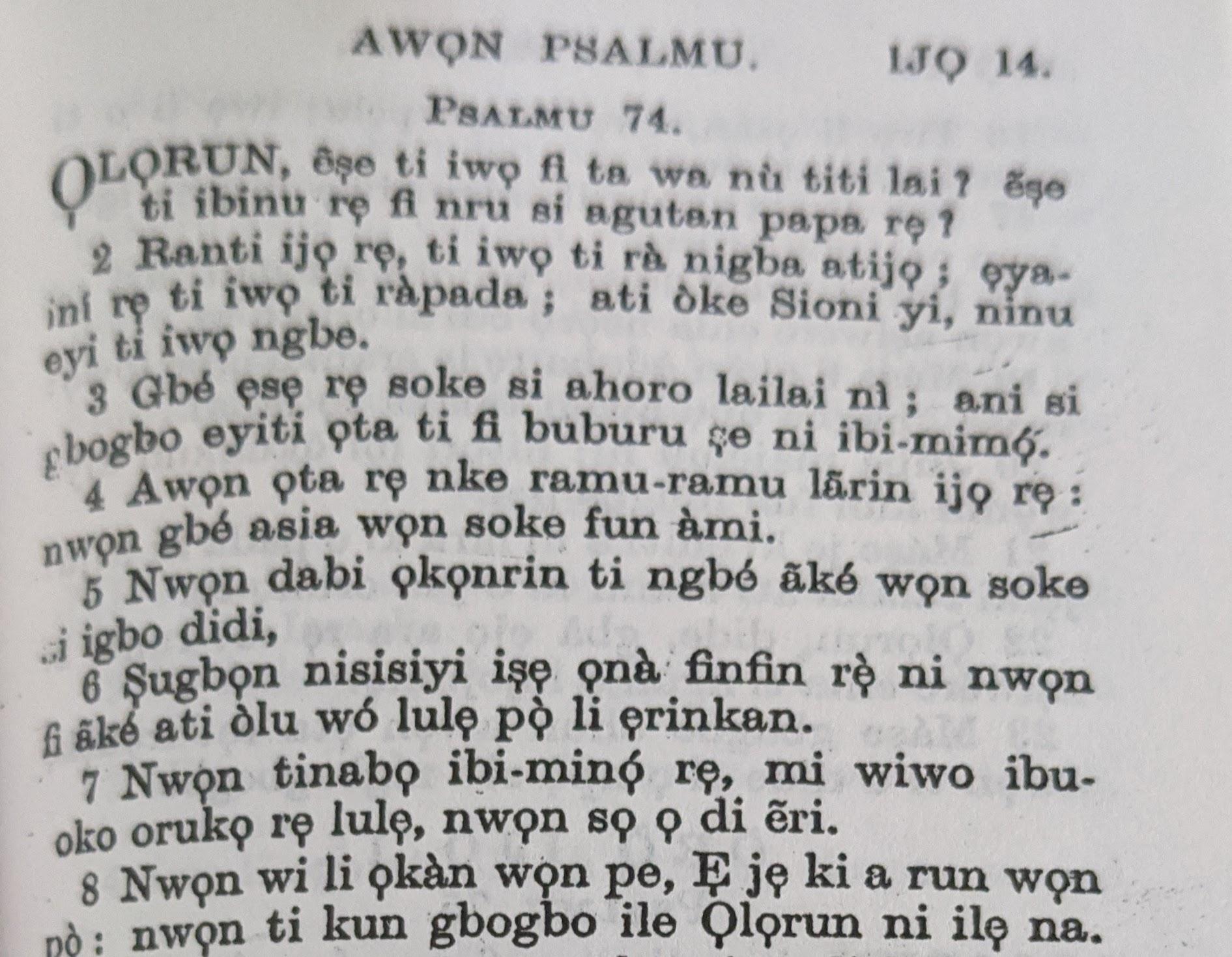}

\textbf{Figure 1:} \emph{Ìwé Àdúrà Yorùbá}. (Yorùbá Church Prayerbook)
Church Missionary Society (1878), showing earlier Yorùbá orthography and
different front and tone-diacritic accommodations.

Over the years, with a wider usage of Yorùbá in writing, a more organic
and naturally less uniform system grew through the language community.
Pamphlets, scripture, songbooks, primers, and other writings in Yorùbá
each carried often similar but usually different systems of writing, but
trends emerged. In 1965, the other major revision to the orthography
happened, through the work of Ayọ̀ Bámgbóṣé, who put forward a number of
substantive changes to the way the language is written, some of which
have been adopted and are present in the way the language is written
today. Four relevant items from the summarised rules are worth noting
here:

5. The tilde to be replaced by double vowels, e.g. {\emph{õrun}
``sun'' to be spelt as \emph{oòrùn}.}

\begin{quote}
{6. The syllabic nasal to be spelt n in all cases, e.g. \emph{ó ḿbọ̀}
``he is coming'' to be spelt \emph{ó ńbọ̀}; \emph{ng ò mọ̀} ``I don't
know'' to be spelt as \emph{n̄ ò mọ̀}.}

{8. Double letters to indicate a consonant, and the spelling
\emph{sh} should be discontinued, e.g. \emph{Ìddó} to be spelt
``Ìdó''\emph{,} \emph{Òshogbo} to be spelt ``Òṣogbo''.}

{9. Tones to be indicated in all cases. The mid tone to be indicated
with a macron only on a syllabic nasal; and the assimilated low tone to
be indicated with a dot, e.g. \emph{panla} ``stockfish'' to be spelt
\emph{pānla}; lóní ``today'' to be spelt \emph{lo.ní} (Ibid., p.3)}
\end{quote}

(Items 1--4 and 7 omitted here; full list in Bámgbóṣé, 1965, p. 3.)

The presence of these instructions on the placement of tone marks ---
and Bámgbóṣé's own acknowledgment in the introduction to the paper ---
shows that the use of diacritics to mark tone had become commonplace
before 1964 when the paper was first presented as a talk to the Ẹgbẹ́
Ìjìlẹ̀ Yorùbá. We had come a long way from Crowther's consonant clusters
like below:

\begin{quote}
\emph{Ekkuhn} (crying, weeping) written today as `ẹkún'

\emph{Ekkung} (region, leopard) written today as `ẹkùn'

\emph{Ekung} (knee) written today as `ekún'

\emph{Ekuń} (perseverance in sickness).
\end{quote}

And so, at least from 1965, we have had the need to write vowels, and
consonants, with both the subdot (e.g. ẹ, ọ, ṣ) and with the tone
marking diacritics (e.g. é, è, ò, ó, ì, í, à, á, ù, ú), those that
combine them both (e.g. ẹ̀, ẹ́, ọ̀, ọ́), and those with special characters
like the caron (e.g. ǎ, ǒ, ě, ǐ), the macron (m̄, n̄), and the circumflex
(ô, â, {ê, î, û).}\footnote{The caron and circumflex are proposed as orthographic contour markers in Túbọ̀sún (2026).}

{Some of the recommendations from Bámgbóṣé did not eventually catch
on: like the \#9 suggestion to have the assimilated low tone ``be
indicated with a dot''. Others, like the \#5 suggestions to replace the
tilde with geminated vowels, have ensured the disappearance of the tilde
totally in Yorùbá writing. Most of the others, especially suggestions
about word presentation, spacing, and tone application, were adopted and
some later improved on. In 1983, he led a second revision to the
orthography (Bámgbóṣé, 1983), suggesting new ways of spelling
(\emph{ayé}, \emph{ẹyẹ}, and \emph{àyà}, instead of ``aiyé'', ``ẹiyẹ'',
and ``àìyà''), that some earlier spellings used in the bible
(\emph{ọkọ̀nrin}) be changed to more modern spellings `ọkùnrin', and
notably that ``the diacritic mark indicating open vowels should be a
vertical bar (tail) or a dot but never a dash'' \raisebox{-0.3em}{\includegraphics[height=1.6em]{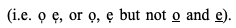}} These, and other recommendations from
Awobuluyi (1978) have mostly defined how Yorùbá is written today.}

\subhead{{The Technology Paradox}}

{Since the advent of computer technology, a new problem has arisen,
which stems from the way marked Yorùbá letters appear as different from
standard Latin-based ones. As shown above, modifications on the standard
Latin scripts were necessary to provide disambiguation and other
orthographic normalization needs. But computer systems, when they were
first introduced to the African market, did not necessarily put these
needs into consideration. I remember using my first computer in 2001 and
realizing that I could not type my full name \textbf{Kọ́láwọlé
Ọlátúbọ̀sún} without having to make sacrifices as to whether I prefered
the diacritics on the vowels or under them, even though both are
necessary.}

{With Microsoft Word, at the time, to write in Yorùbá, one had to go
to the \emph{Insert \textgreater{} Symbol} functionality, which brought
out a box of different symbols from around the world, based on the
Unicode Standard, with each character was given a unique coding to
separate them from one another. And so, to write my name, I would need
to write ``Kolawole Olatubosun" and then replace the needed vowels with
the appropriate symbols I found on the program. Usually, the end result
was one of ``\emph{Kóláwọlé Ọlátúbòsún}'' or ``\emph{Kọláwọlé
Ọlátúbọsún}'', each without the option allowing a combination of
diacritics and subdot on one vowel, resulting in a tonal
misrepresentation of the names.}

{Occasionally, shortcuts were found. There were third-party tone
marking apps that made it possible to write the name as one or the
other, and then apply the last diacritic as an attachment. It was a
temporarily satisfying solution that usually fell apart when the
document was transferred from one platform to the other, maybe from Doc
to PDF. Sometimes, even within the same platform, the document is saved,
and then transferred through a usb drive to another computer maybe for
printing, or sent to a new user for reading, and then instead of the
applied diacritics, one was met with boxes. Sometimes, it was the vowels
that turned into boxes, making the text unreadable. It was worse when
the whole document itself was in Yorùbá, and had taken days and
countless hours to put together through this \emph{Insert \textgreater{}
Symbol} option that was slow and plodding, only for the target reader to
be unable to make sense of the whole document now filled with boxes. It
was a nightmare for class assignments, term papers, and long essays.}

{But that was in the early 2000s, and this related to word processing
applications, a difficulty that would later be classified in Osborn,
Anderson, and Kodama (2008), who, adapting a typology from Taylor
(2000), sorted Latin-based African orthographies into five categories
according to the demands they place on encoding and rendering. It placed
Yorùbá, alongside Igbo, in Category 4: orthographies written in extended
Latin that additionally require combining diacritics. This category, by
their account, introduced difficulties of input, font coverage, and
rendering that the lower categories did not encounter. They identified
diacritic placement as the most persistent of these. Thus what I
experienced in 2001 as a local misfortune of Nigerian computing was in
fact a recognised structural property of the
orthography\textquotesingle s position in the encoding architecture,
described at the Consortium\textquotesingle s own conference nearly two
decades before this paper. It is still unresolved.}

Writing on the internet presented its own challenges. One of the notable
problems I found when I started using Twitter was that the then
140-character count saw new diacritics as separate
characters.\footnote{Twitter doubled this limit to 280 characters in
  November 2017 and rebranded as X in 2023, with an even longer
  character limit, but the diacritic counting problem persisted
  independently of these.} So a name like ``Kọla" would be counted as
four characters while ``Kọ́lá'' was counted as six, because of the two
extra tone marks on the vowels. This was problematic too, as it then
limited the number of words that could be written in Yorùbá on the
platform if one chose to write only in the language. When I led a team
in 2011 to press Twitter to allow the platform to be translated into
Yorùbá\footnote{It Took Only Two Years, But Twitter Is Finally Getting
  Translated Into Yorùbá
  \href{https://techcabal.com/2014/11/14/twitter-yoruba/}{\ul{https://techcabal.com/2014/11/14/twitter-yoruba/}}}{,
this was not one of the issues raised in the campaign, but it should as
well have been. Even if one could use Twitter itself in Yorùbá, the
space to write was still pretty limited because of character counts on
tone markings. This problem, I also found, tied back to Unicode.}

{Font rendering presents a third category of failure, distinct from
the character-count and transfer problems described above.}\footnote{See a longer essay on this problem on the Orisha Image blog, \url{https://www.orishaimage.com/blog/yorubapublishing} (no longer online; archived at \url{https://web.archive.org/web/20190329102140/https://www.orishaimage.com/blog/yorubapublishing}).}

{Finding one's tone markings turning into boxes were problem enough.
In some other cases, the diacritics remained alright, but had moved to a
different vowel than it was on when it was first applied. As a practical
example, seven paragraphs above, where I wrote \raisebox{-0.4em}{\includegraphics[height=1.30em]{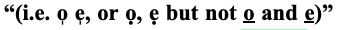}} I
found that if I changed the font 
to EB Garamond, it looks
like this \raisebox{-0.40em}{\includegraphics[height=1.5em]{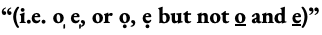}} with
the vertical lines under the first \emph{o} and \emph{e} having moved
from the centre of the vowel to its side, and changing the look and the
characters. This type of problem, I found later, was due to issues with
the font design. But since fonts can only be designed for vowels already
coded by Unicode, I find myself back facing Unicode and its decision
with precomposed as opposed to sequential mark application.}\footnote{In
  a public exchange on Twitter/X in May 2020 (archived at
  https://twitter.com/CharlotteBuff/status/1260251633268686849), a
  Unicode employee confirmed this position, stating that precomposed
  characters for Yorùbá would not be added. The exchange is cited here
  as a documented statement of the Consortium's position rather than as
  a formal publication.} {}

\subhead{{Precomposed Characters and Unicode Normalization}}

{To understand why the problems described have persisted across
decades of digital publishing in Yorùbá, I describe two technical
concepts: precomposed characters and Unicode normalization.}

{A \emph{precomposed character} is a single Unicode code point that
represents a base letter with one or more diacritics already combined
into it. The character \emph{ó} (U+00F3) is a precomposed character: it
is encoded as a single unit, not as the letter \emph{o} (U+006F)
followed by a combining acute accent (U+0301). Precomposed characters
render predictably across platforms, travel correctly between
applications, and are counted as single characters by software that
counts characters. The characters that cause Yorùbá writers the most
difficulty, for instance \emph{ọ́} (open-o with acute accent), \emph{ọ̀}
(open-o with grave accent), \emph{ẹ́} (open-e with acute accent), and
\emph{ẹ̀} (open-e with grave accent), do not exist as precomposed
characters in Unicode. They must be constructed by combining an existing
precomposed character (\emph{ọ}, U+1ECD; or \emph{ẹ}, U+1EB9) with a
separate combining tone mark (U+0301 for acute; U+0300 for grave). This
combining sequence is what causes the rendering failures described
above: that is, the two elements may separate across platforms, the
combining mark may attach to the wrong character, or the sequence may be
counted as two characters rather than one.}

{\emph{Unicode normalization} is the process by which Unicode text is
converted into a canonical form to ensure consistent comparison and
processing. The relevant normalization form for web and document use is
NFC (Canonical Decomposition, followed by Canonical Composition), which
is the form that web browsers, word processors, and most digital
platforms prefer. Under NFC, precomposed characters are kept as single
units, and combining sequences that have a precomposed equivalent are
converted to that equivalent. The critical constraint is this: the NFC
normalization table is frozen. No new precomposed characters can be
added to it (Unicode Consortium, n.d.-a). The combinations \emph{ọ́},
\emph{ọ̀}, \emph{ẹ́}, and \emph{ẹ̀} do not appear in the NFC composition
table and cannot be added to it without a change in Unicode policy.}

The technical reason for their absence is not that such combinations are impossible in principle. Unicode already encodes comparable letters like ệ (U+1EC7) and ộ (U+1ED9) in Vietnamese as single precomposed characters whose canonical decompositions are ẹ (U+1EB9) and ọ (U+1ECD) followed by a combining circumflex; and NFC composes those sequences into single code points. The four Yorùbá characters were simply never encoded, and they cannot usefully be encoded now. The normalization stability policy guarantees that text normalized under one version of Unicode remains normalized under every later version (Unicode Consortium, current version). Any precomposed character added after the composition table was fixed must therefore be listed as a composition exclusion, in the ``post composition version'' category (Davis \& Whistler, current version; Unicode Consortium, 2024, DerivedNormalizationProps.txt), so that NFC decomposes it rather than producing it, and the combining sequence remains the normalized form. For this reason the Consortium does not encode new precomposed Latin letters at all. So, Vietnamese, whose letters were encoded before the table was frozen, received single code points for its stacked diacritics, and Yorùbá, whose equivalent letters were not, cannot. Thus the problem isn't technical impossibility as such, but a gap of timing and policy, which is a consequence of inadequate consideration of the needs of tonal African languages.

\subhead{{Practical Examples}}

{I present here the different ways in which diacritics have been
presented on Yorùbá book covers over the years.}

\begin{center}
\includegraphics[angle=90, height=0.55\textheight,keepaspectratio]{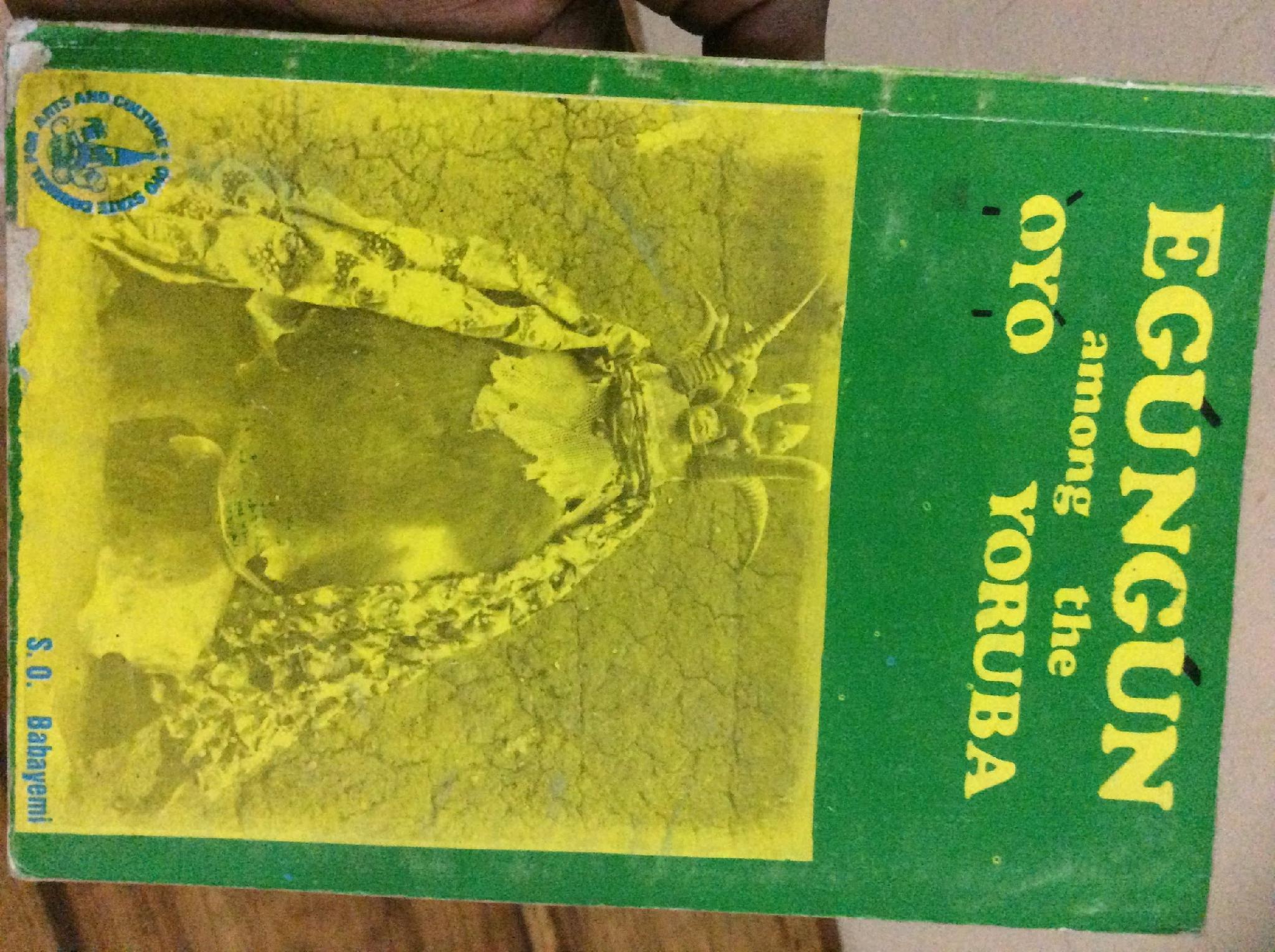}
\end{center}
{\textbf{Figure 2:} Cover of \emph{Egúngún Àmọng the Ọ̀yọ́ Yorùbá}
(1980), showing hand-drawn diacritics applied after typesetting.
\emph{Source: Author's collection.}}

\includegraphics[width=3.94805in,height=4.28646in]{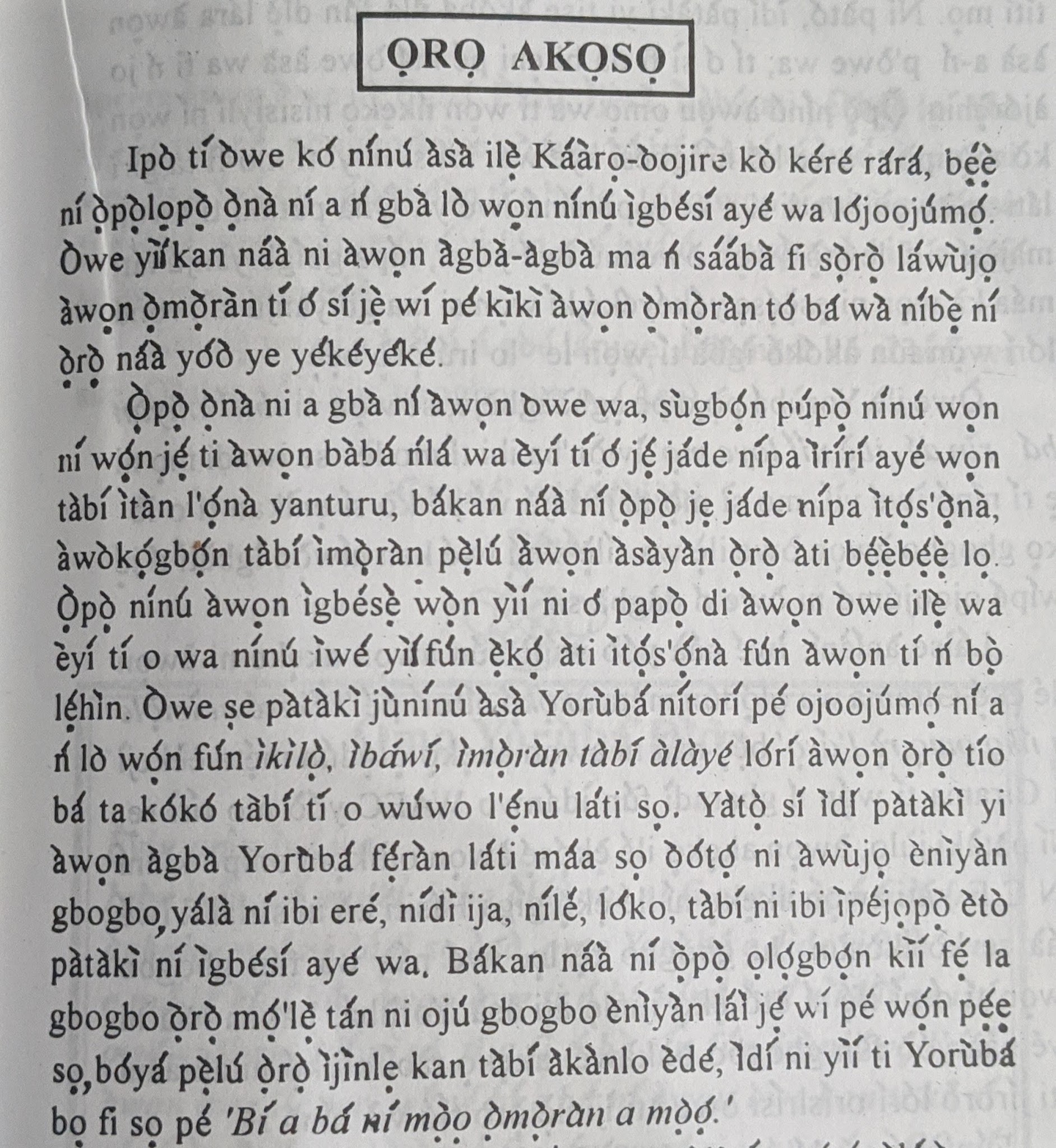}

{F\textbf{igure 3:} A published Yorùbá text, showing hand-drawn
diacritics applied after typesetting.}

{\emph{Source: Àwọn Òwe Àti Àkànlò Èdè Yorùbá (1999) by Gbadé
Aládéòjẹ̀bi} (\emph{Author's collection.})}

\includegraphics[width=3.7456in,height=5.58854in]{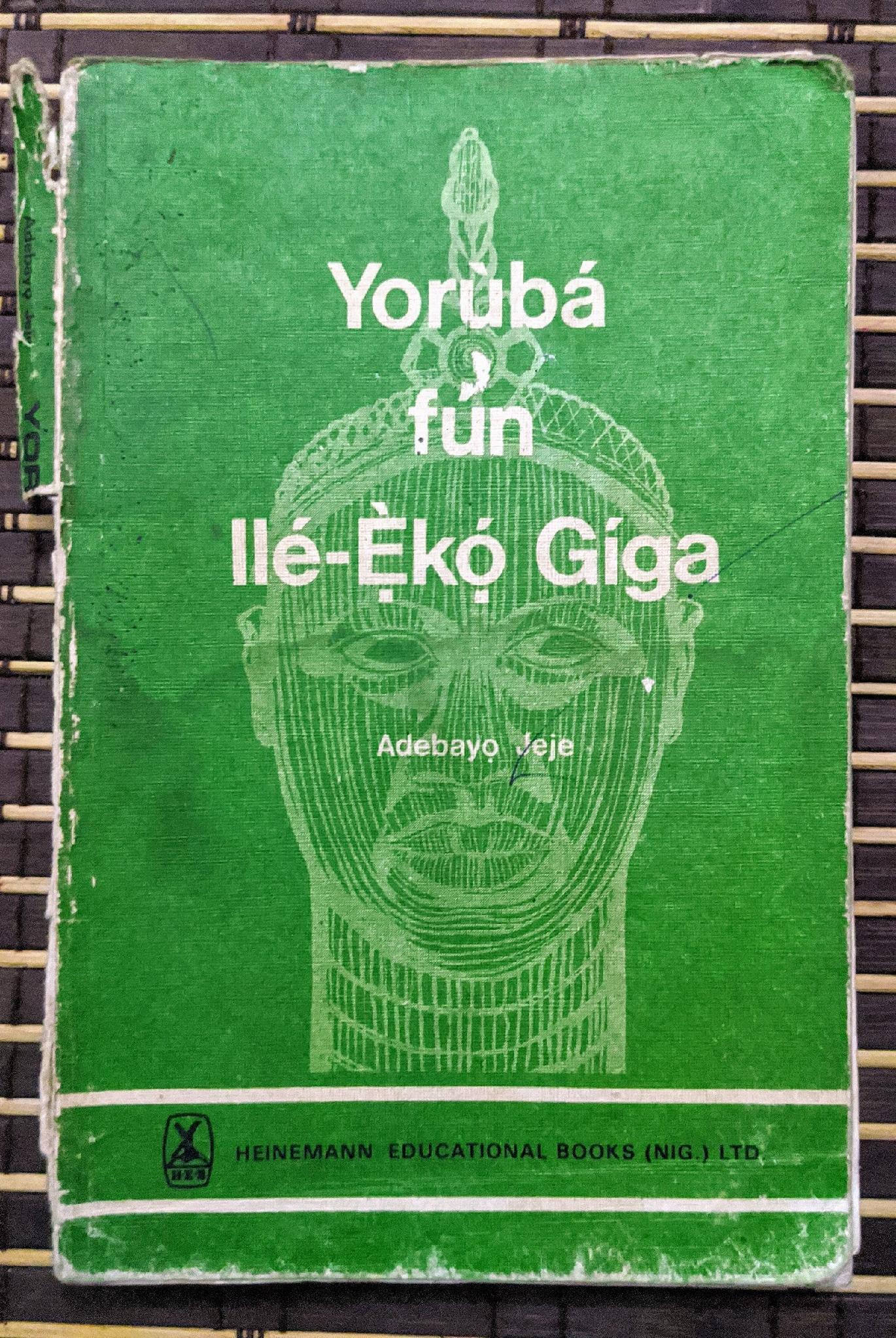}

{\textbf{Figure 4:} Cover of \emph{Yorùbá fún Ilé-Ẹ̀kọ́ Gíga} (1979),
with precomposed vowels correctly rendered. A rarity. \emph{Source:
Author\textquotesingle s collection.}}

\includegraphics[width=4.28507in,height=6.40104in]{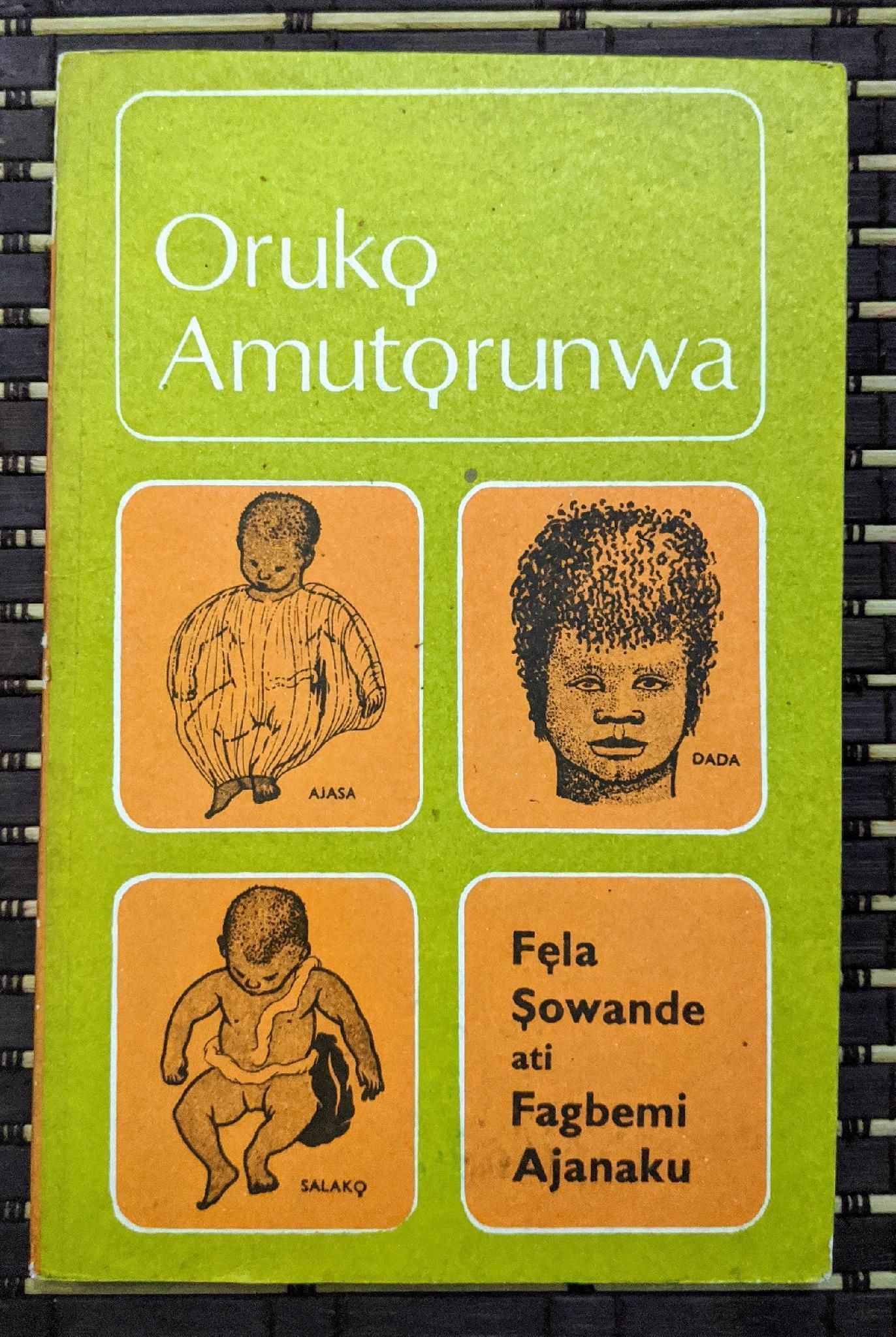}

{\textbf{Figure 5:} Cover of \emph{Orúkọ Àmútọ̀runwá} (1986)
\emph{Source: Author\textquotesingle s collection.}}

  {No need for a combination of subdots and tone marks because none
  of the words and names written there, from the book title to the
  author's name, required it. A serendipitous coincidence.}

{\hfill\break
}\includegraphics[width=3.92153in,height=5.75521in]{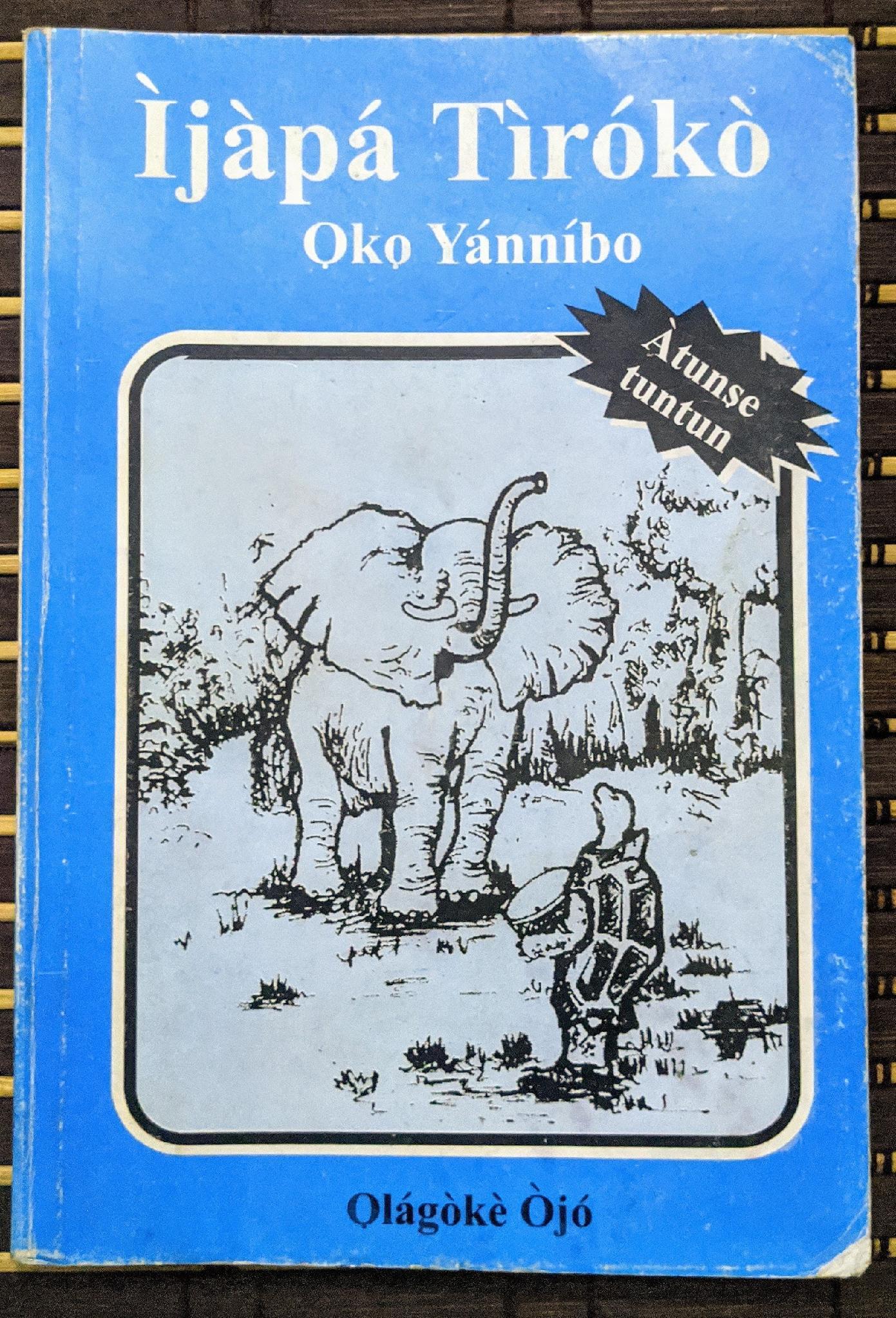}

{\textbf{Figure 6:} Cover of \emph{Ìjàpá Tìrókò} (1973), showing a
title and author name requiring no combination of tone marks and
subdots. \emph{Source: Author\textquotesingle s collection.}}

{All books published in Yorùbá have mostly fallen within these four
options.}

\subhead{{Orthography Meets Online Technology}}

{In September 2019, I began a Chevening Research Fellowship at the
British Library and began to handle books that span the entire length of
the Yorùbá language published history. The oldest book in the Library's
record for Yorùbá is the aforementioned 1843 \emph{Vocabulary} by Samuel
Àjàyí Crowther, meaning my work covered the whole gamut of Yorùbá
language publishing, from the earliest days of its incoherent
orthography to the modern day (Túbọ̀sún, 2022).}
{}

{Because the people in charge of the earliest record creation at the
Library were not native speakers of Yorùbá, and because tonal diacritics
were not always familiar to them from an English-language-speaking
background, they did not always know when a wrong diacritic application
turned an otherwise familiar title into a jumbled mess of floating
symbols, as exemplified by this notorious example entry below:}

\noindent\fbox{\includegraphics[width=\dimexpr\linewidth-2\fboxsep-2\fboxrule\relax]{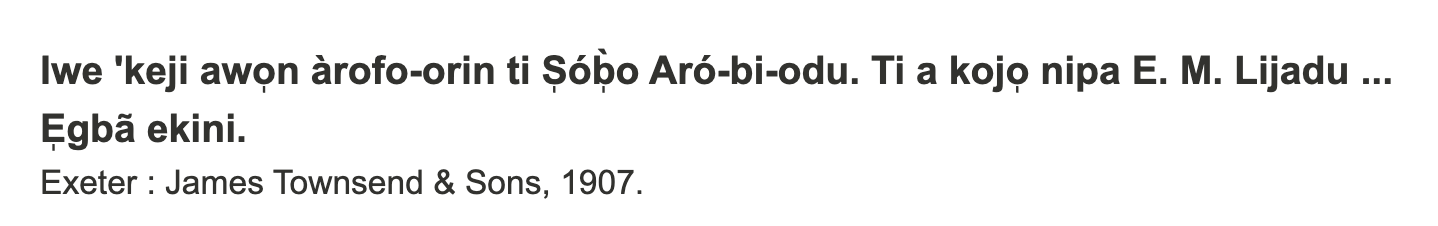}}
\begin{quote}
{\textbf{Figure 7:} British Library catalogue entry showing
diacritics displaced onto consonants rather than vowels in the author
name \emph{Ṣóbọ̀ Aróbíodu}. \emph{Source: British Library catalogue.}}
\end{quote}

\vspace{0.8em}

{In this example, the name of the author, which should have been
written as \emph{Ṣóbọ̀ Aróbíodu} without diacritics hanging around and
under a consonant instead of a vowel.}

{Or the following, where instead of the usual subdot used for Yorùbá
vowels, the ogonek is substituted, perhaps because they look alike:}

\noindent\fbox{\includegraphics[width=\dimexpr\linewidth-2\fboxsep-2\fboxrule\relax]{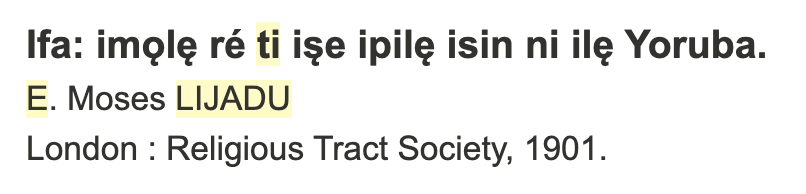}}

{\textbf{Figure 8:} British Library catalogue entry showing the
ogonek substituted for the standard Yorùbá subdot in \emph{imọlẹ},
\emph{iṣẹ}, \emph{ipinlẹ,} and \emph{ilẹ.} \emph{Source: British Library
catalogue.}}

\vspace{0.8em}

{The ogonek under \emph{imọlẹ}, \emph{iṣẹ}, \emph{ipinlẹ}, and
\emph{ilẹ} appear precomposed with the vowels there. That shape is not
used in standard Yorùbá orthography, but it seemed to have sufficed here
for the purpose of cataloguing.}

{In the course of my work at the Library, I came across many more
examples like this, showing improvisational ingenuity by the Library's
cataloguers that would have eventually caused headaches in the Yorùbá
language community. In these two examples, the problem of the absence of
precomposed characters with the subdot \emph{and} the tone marks
together led to problems that could have been avoided.

The cataloguers were not solely to blame. For security reasons, the Library’s cataloguing system is restricted to a set of approved applications, and these were built for well-resourced languages (English, French, German, Arabic and others), with no provision for Yorùbá’s combinations of tone marks and subdots. Faced with characters they could not type, cataloguers improvised with whatever looked closest (Túbọ̀sún, 2022).

The ogonek is not the only substitute. Elsewhere in the same catalogue the
subdot is rendered with a combining cedilla (\emph{\ced{e}k\ced{e}rin} for \emph{ẹkẹrin}) and with a combining vertical line below
(\emph{k\textsyllabic{o}m\textsyllabic{o}nwe},
\emph{\textsyllabic{O}l\textsyllabic{o}run}), the latter a survival of the
older practice of writing the subdot as a short vertical stroke
(Túbọ̀sún, 2022). Each is a distinct code point, and none matches a query
typed with the standard dot below.}

\subhead{{Unicode's Ex/cuse/planation and Its Limitations}}

The Unicode Consortium's official position is that new precomposed Latin
letter combinations will not be added to the standard.\footnote{The
  policy is documented in a public tweet at
  \href{https://twitter.com/CharlotteBuff/status/1260265665879031808?s=20}{\ul{https://twitter.com/CharlotteBuff/status/1260265665879031808?s=20}}}
The FAQ on characters and combining marks states that ``normalization
form NFC (the composed form favored for use on the Web) is frozen --- no
new letter combinations can be added to it'' (Unicode Consortium,
n.d.-a). Unicode Standard Annex \#15 elaborates that ``the Unicode
Consortium strongly discourages new compositions'' and has a ``clear
policy to guarantee the stability of Normalization Forms'' (Davis \&
Whistler, current version). The character encoding stability policies
explain that this constraint is designed to ensure backward
compatibility: once the normalization behaviour of a character is fixed,
it cannot be changed without destabilizing all text encoded in prior
versions of the standard (Unicode Consortium, current version).

In other words, for the Yorùbá writer, write the ọ first, then apply the grave (\textasciigrave) or acute (\textasciiacute) accent later. Or write the ò or ó first, and then apply the underdot.
The latter option was the source of much of my headache as an
undergraduate and later as a research fellow. How do you decide, for
instance, whether the subdot was a dot or a line? In handwritten Yorùbá,
the choice was usually insignificant. But on the computer, where each
was recognized as a different character and treated as such in search
results, it was a much more serious limitation and cause for confusion.

Unicode's stated rationale leaves unanswered why the creation of
genuinely new, discrete code points (not modifications to existing
characters, but new encodings for the specific Yorùbá characters that
have no precomposed equivalents) cannot be accommodated for the small
set of characters the language requires.

Osborn et al. (2008) argue that Latin-based African orthographies are
already covered by the Unicode Standard, and that where support falls
short, the cause lies with fonts and with older operating systems rather
than with the repertoire. But coverage in that sense is a claim about
what can be represented, not about what survives representation. The
failures documented here occur sometimes after the text is correctly
encoded; when it moves between applications, when it is counted, when it
is normalized for comparison, and when it is queried. We do not consider
such a problem fixed when a standard encodes a language but still fails
to render its text stable and findable.

The argument that combining sequences are adequate has been made before,
as Osborn et al. record. In the mid-1990s, the African Language
Resource Council, a joint undertaking of the African Studies Center and
the Linguistic Data Consortium at the University of Pennsylvania,
weighed both an interim eight-bit solution for African extended Latin
and the addition of further precomposed characters to Unicode, but
abandoned the effort on the grounds that support for combining
diacritics was improving. What the forecast did not anticipate is that
the remaining failures would migrate rather than disappear, from glyph
placement to interchange, character counting, normalization, collation,
and search. The precomposition case for African orthographies was
therefore never rejected on its merits, but deferred against an
expectation that has now been partly falsified, and it should be
reopened on that ground.

Unicode has since then, after all, found the resources and the will to
encode tens of thousands of emoji, which are characters, not
combinations of existing ones, since 2010, while the requests from
speakers of tonal African languages with documented orthographic needs
have been ignored. It is, as the W3C's own documentation acknowledges
``a significant I18N problem'' that disproportionately affects minority
and non-European languages (W3C, n.d.).

It is hard to measure how much of this limitation led to the increased
use of non-diacriticised Yorùbá in formal and informal writing over the
years, a phenomenon that McCulloch (2019) documents more broadly as
characteristic of informal digital communication, and which van Esch et
al. (2019) term \textquotesingle online orthographies\textquotesingle{}
in the African language context. The same pressure is visible among
Nigerian writers in English, who frequently omit diacritics from Yorùbá
words embedded in their writing (Túbọ̀sún, 2018b; Túbọ̀sún, 2016b)

Faced with a digital infrastructure that cannot accommodate proper
precomposed diacritics, users often engineer ad hoc phonetic spellings.
By inserting extraneous letters like \textquotesingle h\textquotesingle,
\textquotesingle r\textquotesingle, or
\textquotesingle y\textquotesingle, they attempt to force an
English-based phonetic approximation of Yorùbá vowel length, open
vowels, and tonal contours. In Yorùbá online today, it is not uncommon
to find personal names adapted into forms like the following, taken from
real Facebook profiles:

\begin{itemize}
\item
  \emph{Horlawlahday} (Ọlọ́ládé)
\item
  \emph{Fhamuyiwah} (Fámúyǐwá)
\item
  \emph{Horlarwahley} (Ọláwálé)
\item
  \emph{Ari Ke Adey} (Àríkẹ́ Adé)
\item
  \emph{Harderyemi} (Adéyẹmí)
\item
  \emph{Har Yor Thomih} (Ayọ̌tọ̀mí)
\item
  \emph{Horlalehkan} (Ọlálékan)
\item
  \emph{Adeybaryor} (Adébáyọ̀)
\item
  \emph{Horlatoye} (Ọlátóyè).
\end{itemize}

{The formal spellings are put in brackets, not included in the
original Facebook profile.}

{On Twitter today, the use of `wayrey/wayray' for `wèrè' (\emph{mad
person}), or `shior' for `ṣíọ̀' (\emph{hiss}) or `jor' for `jọ̀ọ́'
(\emph{please}) has become commonplace, to the consternation of those
attached to the formal orthography (though I have argued that many of
these informal adaptations deserve to be understood as resourceful
navigation of a constrained system).} According to van Esch et al.
(2019), some West African language communities have chosen instead to
hand-write their notes, take photos of them, and send these photos to
each other via chat apps, due to the absence of required technology for
their languages. {These do not replace the need for a formal
orthography; thus, the tools to write formally will need to continue to
be available.}

\subhead{{Current Solutions}}

{In 2016, the Yorùbá Names Project released a tone-marking
application for Mac and Windows.}\footnote{A Specially Designed Keyboard
  Allows Yorùbá and Igbo Speakers to Type Their Languages
  \href{https://rising.globalvoices.org/blog/2016/09/26/a-specially-designed-keyboard-allows-yoruba-and-igbo-speakers-to-type-their-languages/}{\ul{https://rising.globalvoices.org/blog/2016/09/26/a-specially-designed-keyboard-allows-yoruba-and-igbo-speakers-to-type-their-languages/}}}
{This works within the current limitations for diacritic
precomposition. Users would have to write the vowel with the subdot
(e.g. \emph{ọ}) and then add the tone mark on top (e.g. \emph{ọ́} or
\emph{ọ̀}) using keystroke combinations on their computers. It has proven
easy to use, and has been downloaded and shared hundreds of times within
the Yorùbá writing community. But the problem of transferability has
remained. When, in 2018, I wrote a journalistic piece for the Guardian
in Nigeria over the demolition of a National Monument}\footnote{The
  three-part journalistic series I published in The Guardian can also be
  found on my blog, starting here:
  \href{http://www.ktravula.com/2016/09/demolishing-history/}{\ul{http://www.ktravula.com/2016/09/demolishing-history/}}}{,
I was surprised and aghast to find that all the tonemarked vowels in the
piece, published in the print version of the paper, had turned into
unrecognizable characters and boxes. The version I published on my
Wordpress blog had not suffered the same fate, although I had used the
same tonemarking software.}

{In 2018, Google unveiled its Gboard keyboard input application. (van
Esch et al. 2019). When I worked at Google in mid-2016, I helped
contribute my Yorùbá language competence to the creation of the baseline
of ``precomposed'' vowels used for Yorùbá and other Nigerian languages
on the platform. It was not exhaustive, alas, so future updates will
need to expand on this initial contribution. I use ``precomposed'' in
quote marks because even though the vowels appear precomposed on mobile
phones, they actually aren't according to Unicode's standard, and often
suffer the same issues earlier listed on internet platforms --- the
problem of counting diacritics as characters, for instance. I have not
seen its characters change to boxes, however, but that could be because
I have never used the Gboard keyboard to write longer essays.}

{Other keyboard applications, such as SwiftKey, also attempt to
support Yorùbá. While I have not personally tested this software,
technical consensus suggests it relies on the same underlying system as
Gboard, depending on non-precomposed characters to render tone-marked
vowels.}

Contour-marked open vowels (ọ̌, ộ, ẹ̌, ệ), which could not previously be typed with any keyboard tool, are now available through the WriteYoruba extension described in Túbọ̀sún (2026). 

\includegraphics[width=5.70833in,height=2.08333in]{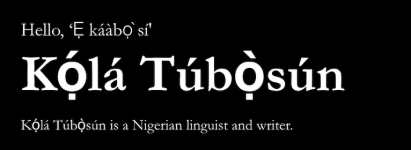}

\textbf{Figure 9:} A screenshot from the author's personal website,
showing font anomalies resulting from diacritic use.

\subhead{The Search Problem}

{The absence of relevant solutions to this issue has contributed to
the frustration of users of the language on the internet, and it's not
limited to Yorùbá. Wrong diacritics on book covers are a discourtesy to
the language and an irritation to the reader. But when found in an
electronic index, they're a barrier to access. The same failure operates
wherever Yorùbá text is matched rather than merely displayed: like in
academic databases and abstracting services, in the author-name fields
of citation indexes and researcher identifier systems, in the archives
of the platforms on which the language is now most actively written. A
scholar publishing under a correctly tone-marked name is liable to
accrue that name inconsistently across indexes, some of which store the
combining sequence, some a stripped form, some a mangled one, and none
of which reliably recognise these as the same person.}

{There is a final consequence of all this. The British Library
catalogue entries reproduced above as Figures 7 and 8 were offered there
as instances of orthographic corruption, where a title queried is
unfindable because of a wrong diacritic. A reader who queries the
author\textquotesingle s name in its correct form, \emph{Ṣóbọ̀ Aróbíodu},
will not retrieve the 1907 record, because the string held in the
catalogue is not that name, and a reader who queries \emph{imọlẹ} with
the standard subdot will not retrieve the 1901 Lijadu record, because
the character sitting in that field is an ogonek and no equivalence rule
relates the two. This problem also exists on the internet, where search
results are specific to the exact query form used. When I search for
``Kọ́lá Túbọ̀sún'', for instance, only results with the exact match of
diacritics are returned. I get different results when I search instead
for ``Kola Tubosun".}

{Current work with Automatic Diacritic Restoration/Application (Orife
et al., 2020) has also begun to focus on a different way of navigating
many of these problems: having automatic application of diacritics by
artificial intelligence and predictive input. This won't solve all the
current problems Yorùbá language writing is facing, but it might make
life slightly easier for users of the language on the computer, or on
the internet.}

\subhead{Conclusion}

Current typing tools, like the WriteYoruba keyboard, the Gboard
implementation, and emerging diacritic restoration systems (Orife et
al., 2020), have substantially eased the problem that Yorùbá writers
faced in the early 2000s. The situation is markedly better than it was.
But the structural gap identified in this paper still subsists. Yorùbá
writers (or people trying to publish in the language) still cannot
produce fully precomposed tonal characters in a form that travels
reliably across all platforms without significant hurdles. And the
search problem, i.e. the failure of diacritic-sensitive search to return
consistent results, remains entirely unresolved.

This situation is not inevitable if specific decisions by the Unicode
Consortium, which are regularly defended on technical grounds, can be
revisited. They have distributional consequences, like disadvantaging
languages that require stacked diacritics while accommodating languages,
and symbol systems like emoji, whose encoding needs are more
straightforwardly met within the existing framework. The W3C has
acknowledged this as a significant internationalization problem (W3C,
n.d.). The Unicode Consortium has the technical capacity and the
institutional authority to address it.

The most direct path forward is a formal Unicode proposal for the
precomposed encodings of the four core Yorùbá characters currently
unrepresented: \emph{ọ̀} (open-o with grave), \emph{ọ́} (open-o with
acute), \emph{ẹ̀} (open-e with grave), and \emph{ẹ́} (open-e with
acute).\footnote{A companion paper (Túbọ̀sún, \emph{2026})
  proposes the caron and circumflex as orthographic contour markers for
  additional characters not yet encoded; the precomposition argument
  applies equally to those forms.} Future work will require a
documentation of the language community's need, and evidence of the
failure of combining sequences to meet that need consistently across
platforms, as well as a sustained engagement with the Unicode Technical
Committee.

This paper, alongside the documented examples of rendering failure it
presents, is intended as a contribution to that evidence base. Font
design, search normalization, and automatic diacritic restoration are
parallel tracks of work that the Yorùbá language technology community is
actively pursuing. A precomposed encoding solution from Unicode remains
the most durable fix which will benefit not only Yorùbá but the many
other tonal African languages whose writers face the same structural
problem.

\vspace{1.5em}

\emph{\textbf{{References}}}

{Àjàyí, J.F. Adé. (1960). ``How Yoruba was Reduced to Writing''.
\emph{Odu: A Journal of Yoruba, Ẹdo and Related Studies}, (8): 49--58.}

{Akinlabí, A. (2004). The sound system of Yorùbá. In N. S. Lawal, M.
N. O. Sadiku, \& A. Dọ̀pámú (Eds.), \emph{Understanding Yoruba life and
culture} (pp. 453--468). Africa World Press.}

{Awóbùlúyì, O. (1978). \emph{Essentials of Yoruba grammar}. Oxford
University Press.}

{Bámgbóṣé, A. (1965). \emph{Yoruba orthography: a linguistic
appraisal with suggestions for reform}. Ibadan University Press.}

{Bámgbóṣé, A. (1966). \emph{A grammar of Yoruba}. Cambridge
University Press.}

{Bámgbóṣé, A. (1983). \emph{Yoruba orthography}. Ibadan University
Press.}

{Brandon, George Edward. (2008). ``From Oral to Digital: Rethinking
the Transmission of Tradition in Yorùbá Religion.'' In \emph{Orisha
Devotion as World Religion}, edited by Jacob K. Olúpọ̀nà and Terry Rey,
448-469. Madison: University of Wisconsin Press.}

{Crowther, Samuel} Àjàyí{. (1843). \emph{A Vocabulary of the
Yoruba Language: Part I English and Yoruba; Part II Yoruba and English:
to which Are Prefixed Grammatical Elements of the Yoruba Language}.
London: Church Missionary Society.}

{Davis, M., \& Whistler, K. (current version). Unicode normalization
forms (Unicode Standard Annex \#15). Unicode Consortium.}
\href{https://unicode.org/reports/tr15/}{{\url{https://unicode.org/reports/tr15/}}}

{van Esch, D., Sarbar, E., Lucassen, T., O'Brien, J., Breiner, T.,
Prasad, M., Crew, E., Nguyen, C., \& Beaufays, F. (2019). Writing across
the world's languages: Deep internationalization for Gboard, the Google
Keyboard. \emph{arXiv preprint arXiv:1912.01218}.}
\href{https://arxiv.org/abs/1912.01218}{{\ul{https://arxiv.org/abs/1912.01218}}}

{McCulloch, G. (2019). \emph{Because Internet: Understanding the New
Rules of Language}. Penguin.}

{Ògúnbíyǐ, Isaac Adéjọjú. (2003). The search for a Yoruba orthography
since the 1840s: Obstacles to the choice of the Arabic script.
\emph{Sudanic Africa}, 14, 77--102.}

{Ọlátúbọ̀sún, K. (2012). Studies of initial tonal acquisition by
American English speakers learning Yoruba {[}Master\textquotesingle s
thesis, Southern Illinois University Edwardsville{]}. Zenodo.}
{\ul{\mbox{\url{https://zenodo.org/records/20833023}}}}

{Orife, I., Adélaní, D., Fasubaa, T., Williamson, V., Oyěwùsì, W.,
Wahab, O., \& Túbọ̀sún, K. (2020). Improving Yorùbá diacritic
restoration. \emph{arXiv preprint arXiv:2003.12855}.}
\href{https://arxiv.org/abs/2003.12855}{{\ul{https://arxiv.org/abs/2003.12855}}}

{Osborn, D., Anderson, D., \& Kodama, S. (2008). Support for modern
African languages and scripts in Unicode/ISO 10646: Where are we today?
Paper presented at the 32nd Internationalization and Unicode Conference,
San Jose, California, September 10, 2008.}

{Oyěníyì, Bùkọ́lá. (2016). ``Oral Tradition'' in \emph{Encyclopedia of
the Yorùbá}, edited by Toyin Falola and Akintunde Akinyemi, 255-256.
Indiana University Press.}

{Pulleyblank, Douglas. (2004). A note on tonal markedness in Yoruba.
\emph{Phonology}, 21(3), 409--425.
\mbox{\href{https://doi.org/10.1017/S0952675704000326}{\ul{https://doi.org/10.1017/S0952675704000326}}}}

{Taylor, C. (2000). \emph{Typesetting African languages: An
investigation} {[}Self-published report{]}. Internet Archive.}
{{\url{https://archive.org/details/TypesettingAfricanLanguages}}}

{Túbọ̀sún, K. (2016a). WriteYoruba: A keyboard input system for Yorùbá
and Igbo. YorubaName.com.}
\href{https://writeyoruba.com}{{\ul{https://writeyoruba.com}}}

{Túbọ̀sún, K. (2016b, August 22). A diligent retelling: Reading Teju
Cole\textquotesingle s essay collection. \emph{ktravula}.}
\href{https://www.ktravula.com/2016/08/a-diligent-retelling-reading-teju-coles-essay-collection/}{{\ul{https://www.ktravula.com/2016/08/a-diligent-retelling-reading-teju-coles-essay-collection/}}}

{Túbọ̀sún, K. (2018a). ``The Web Alternative, Dimensions of Literacy,
and Newer Prospects for African Languages in Today's World.'' In
\emph{Languages, Worlds and Action}. Linguapax International.} {{\href{https://www.linguapax.org/wp-content/uploads/2020/02/Linguapax_Review_2018_revisio\%CC\%81MC-complet-1.pdf}{\ul{https://www.linguapax.org/wp-content/uploads/2020/02/Linguapax\_Review\_2018\_revisio\%CC\%81MC-complet-1.pdf}}}}

{Túbọ̀sún, K. (2018b, April 17). On African languages and literature:
Lessons from Korea {[}Talk transcript{]}. \emph{Agbowó}.}
\href{https://agbowo.org/2018/04/17/on-african-languages-and-literature-lessons-from-korea-kola-tubosun/}{{\ul{https://agbowo.org/2018/04/17/on-african-languages-and-literature-lessons-from-korea-kola-tubosun/}}}

{Túbọ̀sún, K. (2020, September 2). Yorùbá orthography from Àjàyí
Crowther to date: Through the collection items at the British Library:
Progress and problems {[}Conference presentation{]}.
\textquotesingle How Should We Write Yorùbá?\textquotesingle{} British
Library Webinar, London.}
\href{https://www.youtube.com/watch?v=HDn2ou7umxs}{{\ul{https://www.youtube.com/watch?v=HDn2ou7umxs}}}

{Túbọ̀sún, K. (2022). An overview of the British Library Yorùbá
language collection. \emph{Africa Bibliography, Research and
Documentation}, \emph{1}, 47--62.}
\href{https://doi.org/10.1017/abd.2022.3}{{\ul{https://doi.org/10.1017/abd.2022.3}}}

Túbọ̀sún, K. (2026). Marking contour tones in Yorùbá: A typographic and computational proposal. arXiv preprint arXiv:2609.38627. \href{https://arxiv.org/abs/2609.38627}{{\ul{https://arxiv.org/abs/2609.38627}}}

{Túbọ̀sún, K., Olúòkun, A., Adéwuyì, H., \& Adérẹ̀mí, D. (2026). A
situational speech synthesizer for Yorùbá: System design, phonological
rule architecture, and orthographic extensions for contour tones. arXiv
preprint arXiv:2607.18317.}
\href{https://arxiv.org/abs/2607.18317}{{\ul{https://arxiv.org/abs/2607.18317}}}

{Unicode Consortium. (n.d.-a). FAQ: Characters and combining marks. {\url{https://unicode.org/faq/char\_combmark.html}}}

{Unicode Consortium. (current version). Unicode character encoding
stability policies.}
\href{https://www.unicode.org/policies/stability_policy.html}{{\ul{https://www.unicode.org/policies/stability\_policy.html}}}

{Unicode Consortium. (2024). DerivedNormalizationProps.txt.}
{{\url{https://www.unicode.org/Public/UCD/latest/ucd/DerivedNormalizationProps.txt}}}

{W3C. (n.d.). I18N/CanonicalNormalizationIssues.

\mbox{\href{https://www.w3.org/wiki/I18N/CanonicalNormalizationIssues}{\ul{https://www.w3.org/wiki/I18N/CanonicalNormalizationIssues}}}}

{Yip, M. (2002). \emph{Tone}. Cambridge University Press.}

\end{document}